\documentclass[11pt,a4paper]{article}
\usepackage[hyperref]{emnlp2020}
\usepackage{times}
\usepackage{graphicx}
\usepackage{bbm}
\usepackage{latexsym}

\usepackage{subcaption}
\usepackage{stmaryrd}
\usepackage{mathtools}
\usepackage{float}
\usepackage{hyperref}
\usepackage{microtype}
\usepackage[ruled,vlined, noend]{algorithm2e}
\usepackage{amsmath}
\usepackage{amssymb}
\usepackage{ctable}
\usepackage{enumitem}
\usepackage{pifont}
\usepackage{makecell}
\usepackage{url}
\usepackage{footnote}
\usepackage{tablefootnote}

\makesavenoteenv{tabular}
\makesavenoteenv{table}
\usepackage{txfonts}
\aclfinalcopy 

\newcommand{\spider}{\textsc{Spider}}
\newcommand{\sparc}{\textsc{SParC}}
\newcommand{\cosql}{\textsc{CoSQL}}

\title{Semantic Evaluation for Text-to-SQL with Distilled Test Suites}

\author{
Ruiqi Zhong$^{*}$ \quad Tao Yu$^{\dagger}$ \quad Dan Klein$^{*}$ \\
$^{*}$ Computer Science Division, University of California, Berkeley \\
$^{\dagger}$ Department of Computer Science, Yale University \\
ruiqi-zhong@berkeley.edu \quad tao.yu@yale.edu \quad klein@berkeley.edu
}

\date{}

\begin{document}
\maketitle
\begin{abstract}
We propose \textit{test suite accuracy} to approximate semantic accuracy for Text-to-SQL models.
Our method distills a small test suite of databases that achieves high code coverage for the gold query from a large number of randomly generated databases.
At evaluation time, it computes the denotation accuracy of the predicted queries on the distilled test suite, hence calculating a tight upper-bound for semantic accuracy efficiently.
We use our proposed method to evaluate 21 models submitted to the \spider{} leader board and manually verify that our method is always correct on 100 examples.
In contrast, the current \spider{} metric leads to
a 2.5\% false negative rate on average and 8.1\% in the worst case, indicating that test suite accuracy is needed. 
Our implementation, along with distilled test suites for eleven Text-to-SQL datasets, is publicly available.\footnote{Metric implementation and test suites available \href{https://github.com/ruiqi-zhong/TestSuiteEval}{here}, for datasets: \spider{}, \cosql{}, \sparc{}, Academic, Advising, ATIS, Geography, IMDB, Restaurants, Scholar and Yelp.}
\end{abstract}

\section{Introduction}
\begin{figure}[t]
    \centering
    \includegraphics[scale=0.39]{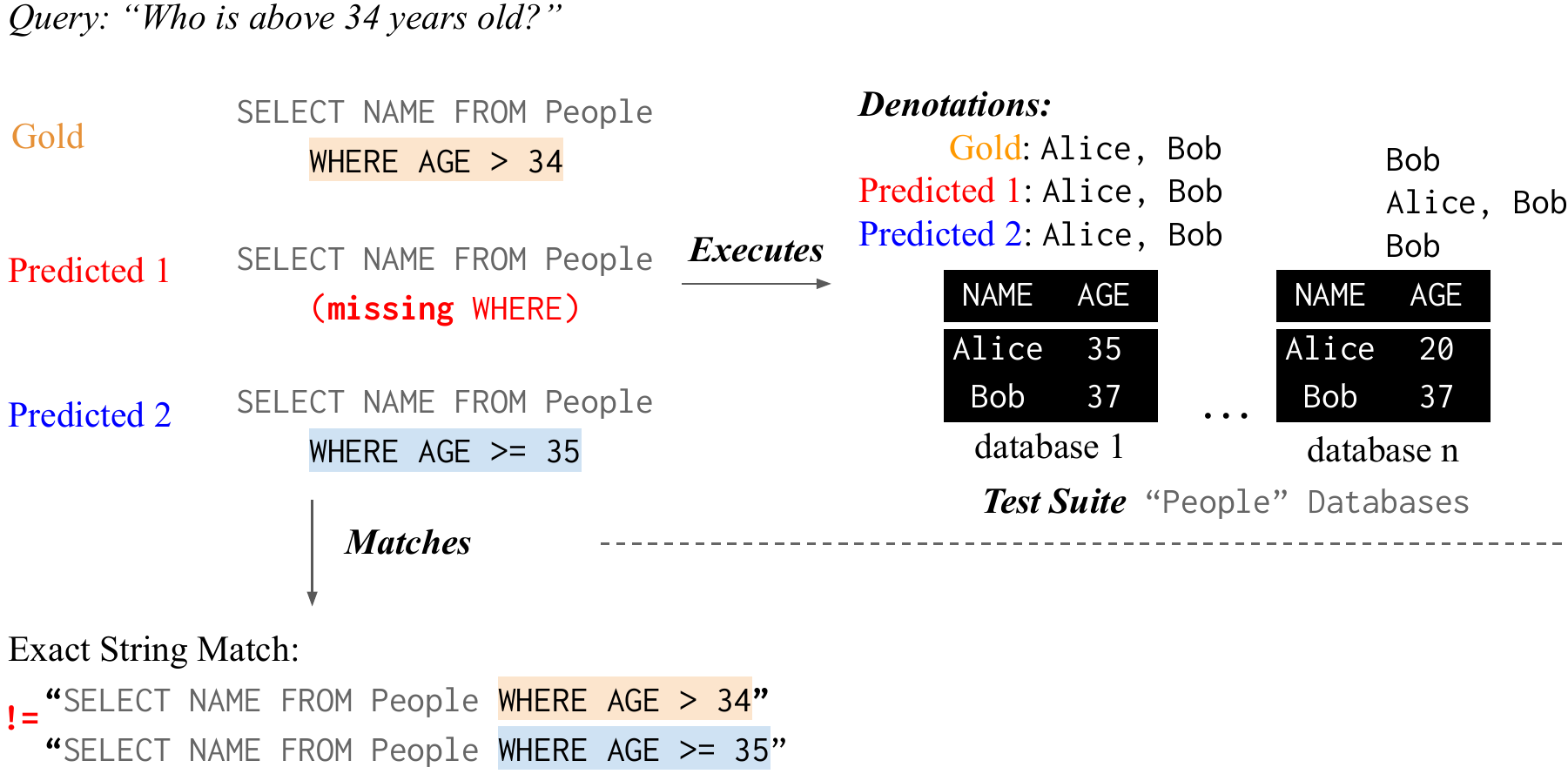}
    \caption{Prediction 2 is semantically correct, and Prediction 1 is wrong. 
    Exact string match judges prediction 2 to be wrong, which leads to \textit{false negatives}. 
    Only comparing denotations on database 1 judges prediction 1 to be correct, which leads to \textit{false positives}.
    Test suite evaluation compares denotations on a set of databases and reduces false positives. 
    }
    \label{fig:semantic-acc-needed}
\end{figure}

A Text-to-SQL model translates natural language instructions to SQL queries that can be executed on databases and bridges the gap between expert programmers and non-experts. 
Accordingly, researchers have built a diversity of datasets \cite{dahl-1989-book, iyer-etal-2017-learning, zhongSeq2SQL2017, yu2018spider} and improved model performances \cite{xu2017sqlnet, Suhr:18context, guo2019towards, Bogin2019RepresentingSS, wang2020ratsql}. 
However, evaluating the semantic accuracy of a Text-to-SQL model is a long-standing problem: we want to know whether the predicted SQL query has the same denotation as the gold for every possible database. 
``Single" denotation evaluation executes the predicted SQL query on one database and compares its denotation with that of the gold. 
It might create \textit{false positives}, where a semantically different prediction (Figure \ref{fig:semantic-acc-needed} prediction 1) happens to have the same denotation as the gold, on a particular database.
In contrast, exact string match might produce \textit{false negatives}: Figure \ref{fig:semantic-acc-needed} prediction 2 is semantically equivalent to the gold but differs in logical form. 

The programming language research community developed formal tools to reliably reason about query equivalence for a restricted set of query types. 
They lift SQL queries into other semantic representations such as K-relations \cite{green2007provenance},
UniNomial \cite{chu2017cosette} and U-semiring \cite{chu2018axiomatic}; then they search for an (in)equivalence proof. 
However, these representations cannot express sort operations and float comparisons,
and hence do not support the full range of operations that Text-to-SQL models can use.
We ideally need a method to approximate semantic accuracy reliably without operation constraints. 

If the computational resources were unlimited, we could compare the denotations of the predicted query with those of the gold on a large number of random databases (Section \ref{sec:sampling}), and obtain a tighter upper bound for semantic accuracy than single denotation evaluation.
The software testing literature calls this idea fuzzing \cite{padhye2019semantic, AFL, lemieux2018perffuzz, QuickCheck}.
However, it is undesirable to spend a lot of computational resources every time when we evaluate a Text-to-SQL model.
Instead, we want to check denotation correctness only on a smaller test suite of databases that are more likely to distinguish\footnote{Section \ref{sec:statement} defines that a database distinguishes two queries if their executions lead to different results.} 
\textit{any} wrong model-predicted queries from the gold.


We propose \textit{test suite accuracy} (Section \ref{sec:statement}) to efficiently approximate the semantic accuracy of a Text-to-SQL model, by checking denotations of the predicted queries on a compact test suite of databases with high code coverage.
We introduce how to construct/search for such a test suite without prior information about model-predicted queries.

Our search objective is formally defined through \textit{neighbor queries} (Section \ref{sec:stress}), which are generated by modifying one aspect of the gold query.
For example, prediction 1 in Figure \ref{fig:semantic-acc-needed}  is a neighbor query of the gold, since they differ only by a ``\texttt{WHERE}" clause.
These neighbor queries are usually semantically different from the gold, and if a test suite can distinguish them from the gold, it is likely to distinguish other wrong queries as well.
The latter holds because distinguishing all neighbors from the gold requires executions on these databases to exercise every modified part of the gold query, hence reflecting comprehensive code coverage and high test quality \cite{miller1963systematic, ammannintroduction}.
Hence, we formalize our objective as finding a small test suite that can distinguish all the neighbors (Section \ref{formalobjective}).

We search under this objective by generating a large number of random databases (Section \ref{sec:sampling}) and keeping a small fraction of them that can distinguish the neighbors from the gold (Section \ref{sec:select}). 
We call this set of databases a \textit{distilled test suite}.
While evaluating model-predicted queries, we only check their denotations on the distilled test suite to approximate semantic accuracy efficiently.

\paragraph{Application} We distill a test suite for the \href{https://yale-lily.github.io/spider}{\spider{}} dataset \cite{yu2018spider} (Section \ref{sec:data}) from 1000 random databases, which can distinguish more than 99\% of the neighbor queries.
We use the test suite to evaluate 21 \spider{} leader board submissions, 
randomly sample 100 model-predicted queries where our method disagrees with 
exact set match (ESM, the current \spider{} official metric), and manually verify that our method is correct in \textit{all} these cases (Section \ref{reliability}).


We use test suite accuracy as a proxy for semantic accuracy to examine how well ESM approximates the semantic accuracy (Section \ref{errors}), and identify several concerns.
(1) ESM tends to underestimate model performances, leading to a 2.5\% false negative rate on average and 8.1\% in the worst case.
(2) ESM does not reflect all improvements in semantic accuracy. 
For example, it undervalues a high-score submission with 61\% semantic accuracy by 8\%, but instead favors five other submissions with lower semantic accuracy, thus misrepresenting state of the art.
(3) ESM becomes poorer at approximating semantic accuracy on more complex queries.
Since models are improving and producing harder queries, ESM deviates more from semantic accuracy. 
We need test suite accuracy to better track progress in Text-to-SQL development.

Our main paper focuses on \spider{}. 
However, we also generated distilled test suites for other popular text-to-SQL datasets including \cosql{} \cite{yu-etal-2019-cosql}, \sparc{} \cite{yu-etal-2019-sparc}, Academic \cite{li2014constructing}, Advising \cite{finegan2018improving}, ATIS \cite{dahl-etal-1994-expanding}, Geography \cite{10.5555/1864519.1864543}, IMDB \cite{yaghmazadeh2017sqlizer}, Restaurants \cite{10.1145/604045.604070}, Scholar \cite{iyer-etal-2017-learning} and Yelp \cite{yaghmazadeh2017sqlizer}.
We will release our test suites\footnote{https://github.com/ruiqi-zhong/TestSuiteEval} and the details of these datasets can be seen in Appendix \ref{sec:otherstats}.


To summarize, we contribute:
\begin{itemize}
    \item A method and a software to create compact high quality test suites for Text-to-SQL semantic evaluation.
    \item Test suites to reliably approximate semantic accuracy for eleven popular datasets.
    \item A detailed analysis of why current metrics are poor at approximating semantic accuracy.
\end{itemize}

\section{Problem Statement} \label{sec:statement}
Let $w \in \mathcal{W}$ be a database input to a SQL query $q \in Q$, and $\llbracket q\rrbracket_{w}$ be the denotation of $q$ on $w$,\footnote{As in \citet{yu2018spider}, we use Sqlite3 to obtain the denotation. Define $\llbracket q\rrbracket_{w} = \bot$ if execution does not end, which is implemented as timeout in practice.}  where $\mathcal{W}/Q$ is the space of all databases/SQL queries.
Two queries $q_{1}$ and $q_{2}$ are semantically equivalent if their denotations are the same for all possible databases, i.e.
\begin{equation}
    \forall w\in \mathcal{W}\space, \llbracket q_{1}\rrbracket_{w} = \llbracket q_{2}\rrbracket_{w}
\end{equation}

We refer to the ground truth query as $g$ and the model-predicted query to be evaluated as $q$. 
Ideally, we want to evaluate whether $q$ is semantically equivalent to $g$ (abbreviated as \textbf{semantic accuracy}), which is unfortunately undecidable in general \cite{chu2017cosette}.
Traditionally, people evaluate a model-predicted query $q$ by either \textbf{exact string match} or compare denotations on a single database $w$ (abbreviated as \textbf{single denotation accuracy}).
Exact string match is too strict, as two different strings can have the same semantics. 
Single denotation evaluation is too loose, as the denotations of $g$ and $q$ might be different on another database $w$.

We use \textbf{test suite} to refer to a set of databases.
A database $w$ \textbf{distinguishes} two SQL queries $g, q$ if $\llbracket g\rrbracket_{w} \neq \llbracket q\rrbracket_{w}$, and a test suite $S$ distinguishes them if one of the databases $w \in S$ distinguishes them:
\begin{equation}
    \exists w \in S, \space \llbracket g\rrbracket_{w} \neq \llbracket q\rrbracket_{w}
\end{equation}
For convenience, we define the indicator function:
\begin{equation}
   D_{S}(g, q) \coloneqq \mathbbm{1}[\text{$S$ distinguishes $g, q$}]
\end{equation}
We use the test suite $S$ to evaluate a model-predicted query $q$: $q$ is correct iff $D_{S}(g, q) = 0$;
i.e., $g$ and $q$ have the same denotations on all databases in $S$.

To summarize, if $M_{1} \Rightarrow M_{2}$ means that ``correctness under $M_{1}$ implies correctness under $ M_{2}$", exact match $\Rightarrow$ semantic accuracy $\Rightarrow$ test suite accuracy $\Rightarrow$ single denotation accuracy.
Our goal is to construct a test suite of databases $S$ to obtain a tight upper bound on semantic accuracy with test suite accuracy reliably and efficiently. 

\section{Desiderata}\label{sec:desiderata}
Since we want to construct a test suite $S$ of databases for each gold query $g$, we use $S_{g}$ to denote the target test suite.
Before describing how to generate $S_{g}$, we first list two criteria of a desirable test suite. 
Later we construct $S_{g}$ by optimizing over these two criteria.

\paragraph{Computational Efficiency.}
 We minimize the size of $S_{g}$ to speed up test suite evaluations.
 
 \paragraph{Code Coverage.}
The test suite needs to cover every branch and clause of the gold query such that it can test the use of every crucial clause, variable, and constant. 
For example, database 1 in Figure \ref{fig:semantic-acc-needed} alone does not have a row where ``\texttt{AGE} $\leq$ 34" and hence does not have comprehensive code coverage.

\subsection{Measure Coverage through Neighbors} \label{sec:stress}
\begin{figure}
    \centering
    \includegraphics[scale=0.55]{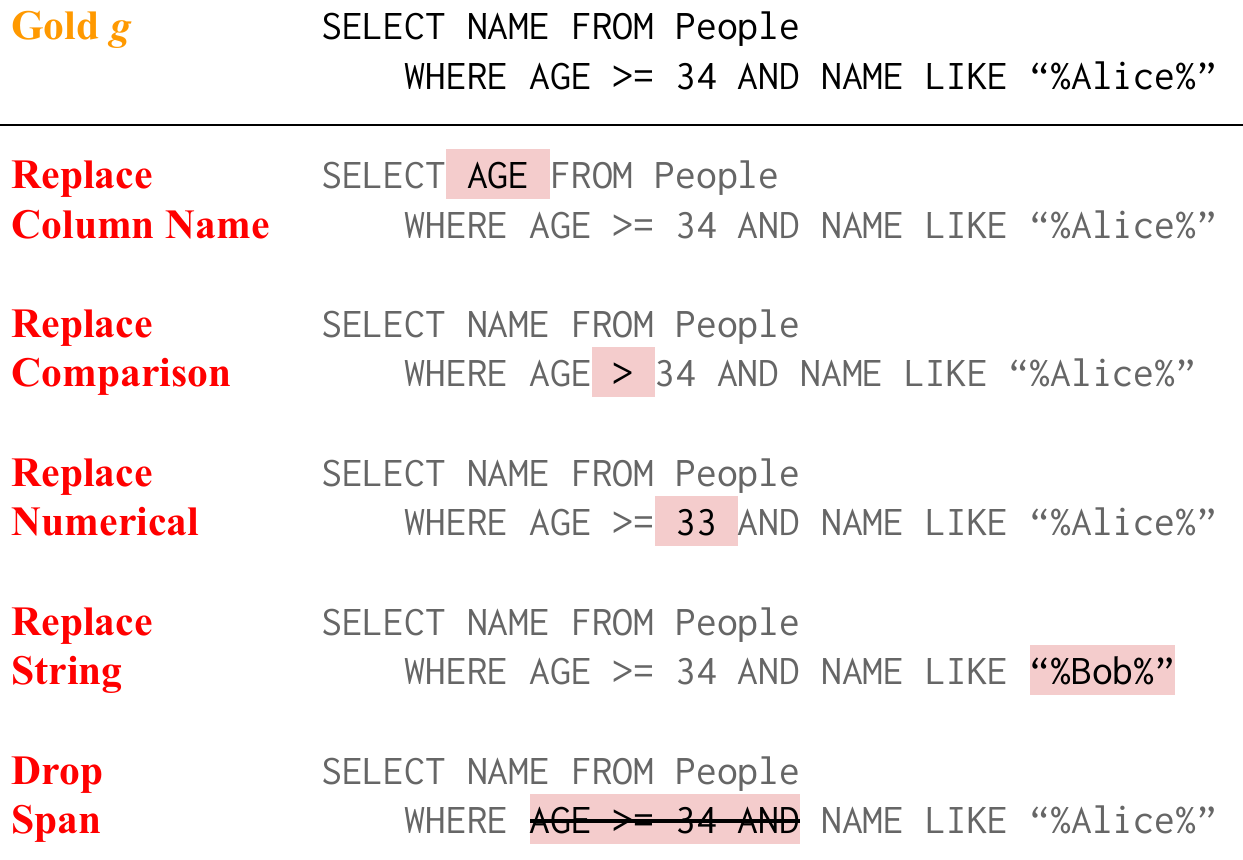}
    \caption{Automatically generating a set of neighbor queries $N_{g}$. 
    We apply one type of modification to the original gold query $g$ at a time. 
    The modified queries are likely to be semantically close but inequivalent to the gold.
    }
    \label{fig:stress}
\end{figure}

We measure the code coverage of a test suite by its ability to distinguish the gold query from its \textit{neighbor queries}: a set of SQL queries that are close to the gold in surface forms but likely to be semantically different.
To generate them, we modify one of the following aspects of the gold query (Figure \ref{fig:stress}): (1) replace an integer (float) constant with either a random integer (float) or its value $\pm$ 1 (0.001); (2) replace a string with a random string, its sub-string or a concatenation of it with another random string; (3) replace a comparison operator/column name with another; (4) drop a query span unless the span does not change the semantics of the query. 
For example, the ``\texttt{ASC}" keyword does not change the semantics because it is the default sort order.
We then remove modified queries that cannot execute without any errors.

Note that our method does not pre-determine the number of neighbor queries.
It is adaptive and generates more neighbors for longer and more complex queries since there are more spans to drop and more constants to replace.

Neighbor queries have two desirable properties.
First, they are likely to be semantically different from the gold query.
For example, ``\texttt{$>$ 34}" is semantically different from ``\texttt{$\geq$ 34}" (replace comparison operator) and ``\texttt{$>$ 35}" (replace constants);
however, we only apply one modification at a time, since ``\texttt{$>$ 34}" is semantically equivalent to ``\texttt{$\geq$ 35}" for an integer.
Secondly, their subtle differences from the gold require the test suite to cover all the branches of the gold query. 
For example, the database needs to have people above, below and equal to age 34 to distinguish all its neighbors.
Hence, the test suite tends to have high quality if it can distinguish the gold from all its neighbors.

We use $N_{g}$ to denote the set of neighbor queries of the gold query $g$.

\subsection{Optimization Objective} \label{formalobjective}
To recap, our objective is to search for a \textit{small} test suite $S_{g}$ that can distinguish as \textit{many} neighbor queries as possible. 
Formally, we optimize over $S_{g}$ with the objective below:
\begin{equation} \label{optmize}
\begin{aligned}
&\text{minimize } |S_{g}|\\
s.t. \quad &\forall q \in N_{g},\space D_{S_{g}(g, q)} = 1
\end{aligned}
\end{equation}

\section{Fuzzing}\label{sec:fuzzing}
We optimize the above objective through fuzzing: a software testing technique that generates a large number of random inputs to test whether a program satisfies the target property (e.g., SQL equivalence).
We describe a procedure to sample a large number of random databases and keep a small fraction of them to distill a test suite $S_{g}$.

\subsection{Sampling Databases} \label{sec:sampling}
\begin{figure}
    \centering
    \includegraphics[scale=0.45]{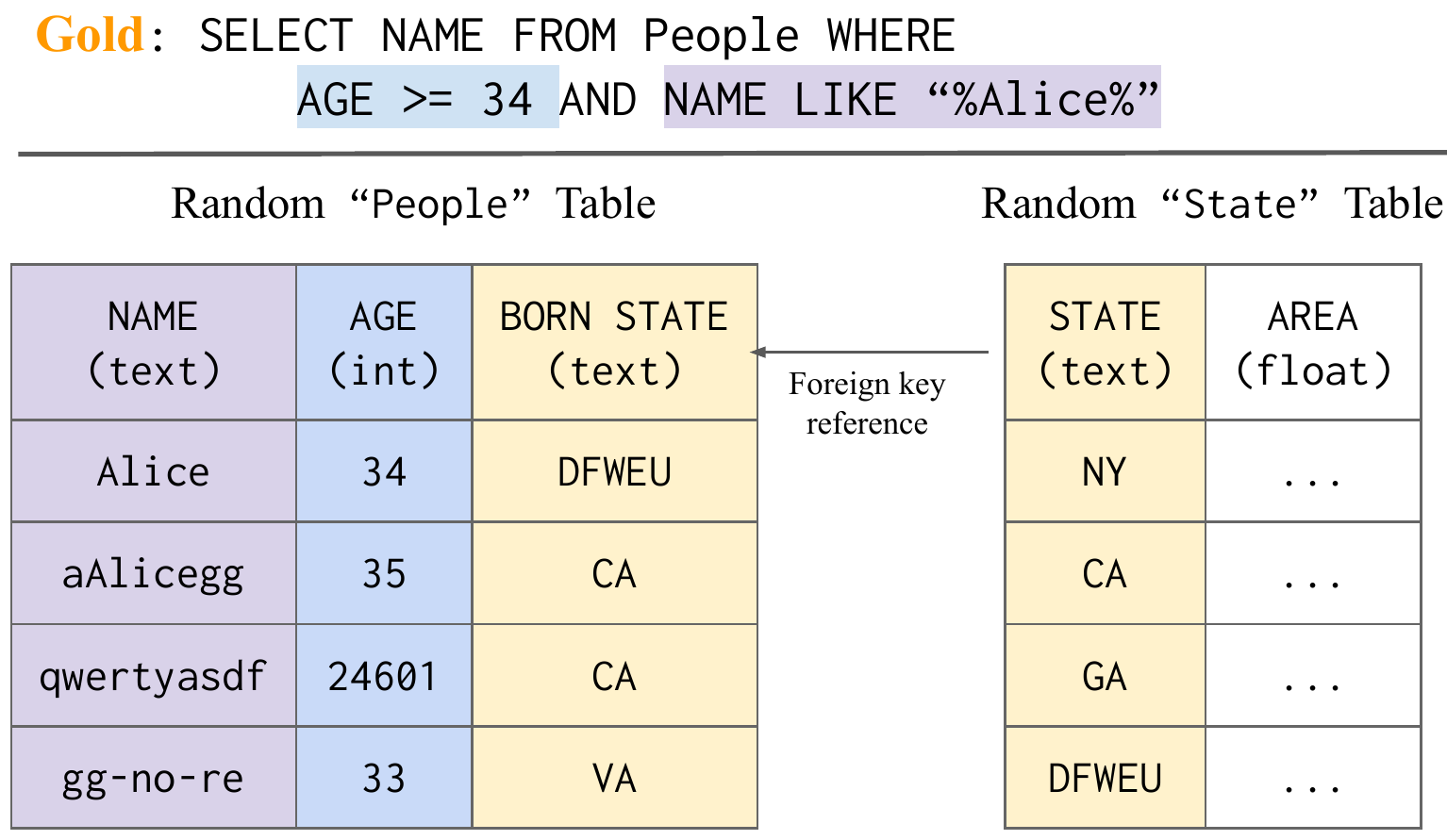}
    \caption{A random database input $w$ from the distribution $\mathcal{I}_{g}$, where $g$ is the gold SQL query. 
    We generate the ``\texttt{State}" column before the ``\texttt{BORN STATE}" column because the latter has to be a subset of the former. 
    Each element of the column ``\texttt{BORN STATE}" is sampled uniformly at random from the parent ``\texttt{STATE}" column.
    For the column that has data type int/string, each element is either a random number/string or a close variant of a constant used the gold query.
    }
    \label{fig:fuzz}
\end{figure}

A database $w$ needs to satisfy the input type constraints of the gold program $g$, which include using specific table/column names, foreign key reference structure, and column data types.
We describe how to generate a random database under these constraints and illustrate it with Figure \ref{fig:fuzz}. 

If a column $c_{1}$ refers to another column $c_{2}$ as its foreign key, all elements in $c_{1}$ must be in $c_{2}$ and we have to generate $c_{2}$ first.
We define a partial order among the tables: table $A <$  table $B$ if $B$ has a foreign key referring to any column in table $A$.
We then generate the content for each table in ascending order found by topological sort.
For example, in Figure \ref{fig:fuzz}, we generate the ``\texttt{State}" table before the ``\texttt{People}" table because the latter refers to the former.

We now sample elements for each column such that they satisfy the type and foreign key constraints. 
If a column $c_{1}$ is referring to another column $c_{2}$, each element in $c_{1}$ is uniformly sampled from $c_{2}$.
Otherwise, if the column is a numerical(string) type, each element is sampled uniformly from $[-2^{63}, 2^{63}]$ (a random string distribution).
We also randomly add in constant values used in $g$ (e.g., 34 and ``Alice") and their close variants (e.g., 35 and ``aAlicegg") to potentially increase code coverage.
We denote the database distribution generated by this procedure as $\mathcal{I}_{g}$.

\subsection{Distilling a Test Suite} \label{sec:select}
We use samples from $\mathcal{I}_{g}$ to construct a small test suite $S_{g}$ such that it can distinguish as many neighbor queries (Section \ref{sec:stress}) in $N_{g}$ as possible.
We initialize $S_{g}$ to be empty and proceed greedily. 
A database $w$ is sampled from the distribution $\mathcal{I}_{g}$; if $w$ can distinguish a neighbor query that cannot be distinguished by any databases in $S_{g}$, we add $w$ to $S_{g}$.
Appendix \ref{algo-description} gives a more rigorous description.
In the actual implementation, we also save the disk space by sharing the same random database $w_{t}$ across all gold SQL queries that are associated with the same schema.
Though this algorithm is far from finding the optimal solution to Objective \ref{optmize}, in practice, we find a test suite that is small enough to distinguish most neighbor queries.

\section{Evaluation Setup} \label{sec:data}
We introduce the dataset and the model-predicted queries we use to study our test suite evaluation. 
We also adapt our test suite evaluation and the official \spider{} metric to make a fair comparison.

\subsection{Dataset}
We generate test suites $S_{g}$ for the development set of \spider{} \cite{yu2018spider}, which contains 1034 English utterances and their corresponding SQL queries, spanning across 20 different database schemata. 
It stratifies data into four categories ($\mathsf{easy}$, $\mathsf{medium}$, $\mathsf{hard}$, and $\mathsf{extra hard}$) according to difficulty level measured by gold SQL complexity.
We decide to focus on \spider{} because it invites researchers to submit their model-predicted queries and requires them to follow a standard format, which makes it convenient to study a wide variety of model-predicted queries.

\subsection{Model-Predicted Queries} 

We obtain the development set model-predicted queries from 21 submissions.\footnote{The model-predicted queries are available \href{https://github.com/ruiqi-zhong/TestSuiteEval/tree/main/predictions}{here}.}
They include models from \citet{guo2019towards, bogin2019representing, choi2020ryansql, wang2020ratsql}.
\footnote{Many dev set submissions do not have public references.}
These models capture a broad diversity of network architectures, decoding strategies, and pre-traning methods, with accuracy ranging from below 40\% to above 70\%.
The first author obtained these model-predicted queries from the second author \textit{after} producing the test suites to ensure that our method is general and not tailored to a specific family of model-predicted queries.

\subsection{Metric Adaptation} \label{adapted-metric}
The \spider{} official evaluation metric is exact set match (abbreviated as ESM) \cite{zhongSeq2SQL2017, yu2018spider}. 
It parses the gold and model-predicted queries into sub-clauses and determines accuracy by checking whether they have the same \textit{set} of clauses.
It improves over exact string matching by preventing false negatives due to semantically equivalent clause reordering. 
However, it is still considered a strict metric and creates false negatives.  

To further reduce false negatives, the actual implementation of the official \spider{} metric is looser.
We list all of its major differences from the standard ESM below;
accordingly, we either adapt our test suite evaluation or fix the \spider{} implementation to make a fair comparison.

(1) The \spider{} metric does not check constant prediction correctness.
Therefore, our adapted test suite evaluation enumerates all possible ways to replace the constants in a model-predicted query with the gold constants and consider a model-predicted query to be correct if one of the replacements passes the test suite.
(2) The \spider{} metric does not check column order, so our adapted evaluation considers two denotations equivalent if they only differ by column order.
(3) The \spider{} evaluation script accidentally ignores any join predicate. 
We fix this bug.
(4) The \spider{} metric does not check table variable names. \citet{yu2018spider} implemented this because different intermediate tables can contain the same column, hence selecting any of them is equivalent. 
We keep this feature since it effectively rules out many false negatives.
However, it also introduces new false positives (e.g., Figure \ref{fig:fpfn-analysis} row 1).

Unless we explicitly specify, in the rest of our paper, ``ESM" and ``test suite accuracy" refer to these adapted metrics rather than the original ones.

\section{Results}
We first show that our test suite evaluation is reliable by verifying that the test suite distinguishes most neighbor queries, and always makes the correct judgement on 100 model-predicted queries we manually examined (Section \ref{reliability}).
Then we use test suite accuracy as a proxy for semantic accuracy to calculate the error rate of the existing commonly used metrics (Section \ref{errors}) and their correlations (Section \ref{sec:metric-corr}).
At last we discuss the computational efficiency of test suite evaluation (Section \ref{sec:efficiency}).

\begin{figure*}
    \centering
    \includegraphics[scale=0.75]{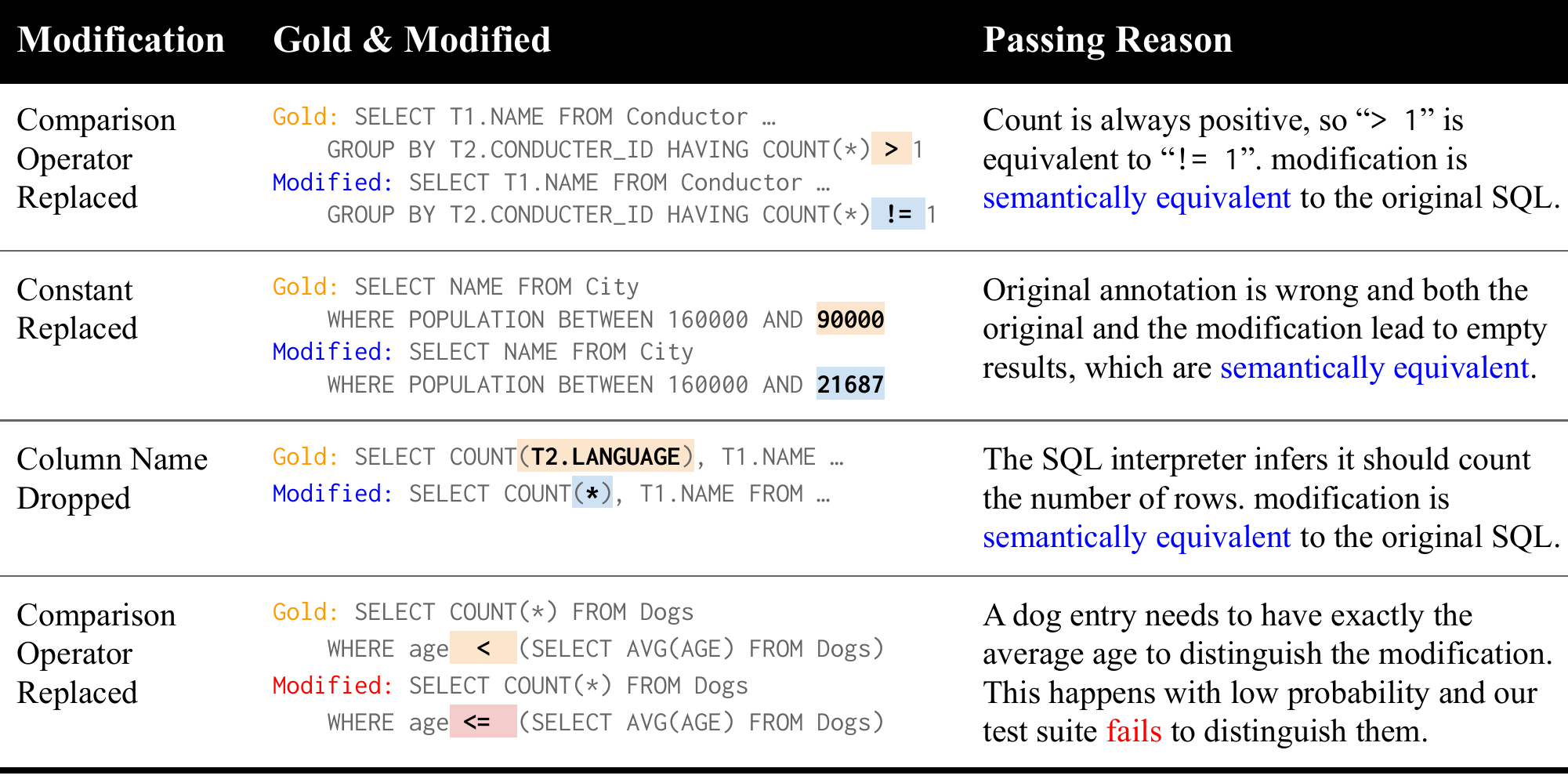}
    \caption{Representative modifications in $N_{g}$ that produce the same results as the gold (pass) on all sampled databases. }
    \label{fig:undisginuisehdst}
    \vspace{-2mm}
\end{figure*}

\begin{figure}[h]
    \centering
    \includegraphics[scale=0.40]{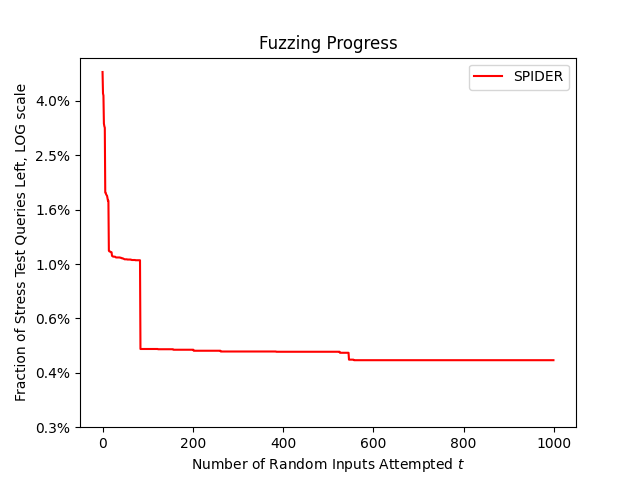}
    \caption{
    The progress of fuzzing (Section \ref{sec:select}).
    The $x$-axis represents the number of random databases attempted ($t$), and the $y$-axis (re-scaled by $\log$) is the fraction of neighbor queries left.
    $y$-value at $x=0$ is the fraction of neighbors left after checking denotations on the database provided by \citet{yu2018spider}.
    }
    \label{fig:fuzz-progress}
    \vspace{-3mm}
\end{figure}

\subsection{Reliability} \label{reliability}
\paragraph{Distinguish Neighbor Queries.}
For each gold query in \spider{}, we generate on average 94 neighbor queries (Figure \ref{fig:stress}). 
We sample 1000 random databases for each database schema and run fuzzing (Section \ref{sec:select}) to construct $S_{g}$, which takes around a week on 16 CPUs.
Figure \ref{fig:fuzz-progress} plots the fraction of neighbor queries that remain undistinguished after attempting $t$ random databases.

Checking single database denotation fails to distinguish $5\%$ of the neighbor queries,
and the curve stops decreasing after around 600 random databases;
1000 random databases can distinguish $> 99\%$ of the neighbor queries.
A large number of random databases is necessary to achieve comprehensive code coverage.

Figure \ref{fig:undisginuisehdst} presents some typical neighbor queries that have the same denotations as the gold on all the databases we sampled.
These queries are only a small fraction (1\%) of all the neighbors; 
in most cases they happen to be semantically equivalent to the gold.
We acknowledge that our fuzzing based approach has trouble distinguishing semantically close queries that differ only at a floating-point precision (e.g. ``\texttt{$\leq$ 2.31}" vs. ``\texttt{$<$ 2.31}").
Fortunately, however, we cannot find a false positive caused by this weakness in our subsequent manual evaluation.

\paragraph{Manual Evaluation.}
Even though our test suite achieves comprehensive code coverage, we still need to make sure that our method does not create any false positive on model-predicted queries.
We focus on the queries from the 21 submissions that are considered incorrect by ESM but correct by our test suite evaluation, randomly sampled and manually examined 100 of them.
\textit{All} of them are semantically equivalent to the gold query;
in other words, we did not observe a single error made by our evaluation method. 
We release these 100 model-predicted queries along with annotated reasons for why they are equivalent to the gold labels,\footnote{Manual examinations are available \href{https://github.com/ruiqi-zhong/TestSuiteEval/blob/main/ESMFalseNegatives.tsv}{here}.} such that the research community can conveniently scrutinize the quality of our evaluation method.

We also confirm that our method can reliably evaluate model-predicted queries on WikiSQL \cite{zhongSeq2SQL2017}.
We refer the readers to Appendix \ref{validateonwikisql} for further experimental details.

\begin{table}[h]
    \centering
    \begin{tabular}{cccc}
        Difficulty & Mean & Std & Max \\
        \hline
        $\mathsf{easy}$ (\%) & 0.5\,/\,2.2 & 0.5\,/\,1.3 & 2.0\,/\,\phantom{0}7.2\\
        $\mathsf{medium}$ (\%) & 0.2\,/\,1.9 & 0.3\,/\,1.9 & 0.7\,/\phantom{0}\,8.0\\
        $\mathsf{hard}$ (\%) & 0.5\,/\,4.4 & 1.2\,/\,3.8 & 4.0\,/\,12.1\\
        $\mathsf{extra}$ (\%) & 1.7\,/\,3.2 & 1.8\,/\,1.6 & 5.3\,/\phantom{0}\,8.2\\
        \hline
        all data (\%) & 0.5\,/\,2.6 & 1.0\,/\,1.7 & 2.0\,/\phantom{0}\,8.1\\
    \end{tabular}
    \caption{The false positive/negative rate of the adapted exact set match metric (Section \ref{adapted-metric}) for each difficulty split. We report the mean / std / max of these two statistics among 21 dev set submissions.}
    \label{tab:fpfn}
\end{table}

\begin{table}[h]
    \centering
    \begin{tabular}{cccc}
        Difficulty & Mean & Std & Max \\
        \hline
        $\mathsf{easy}$ (\%) & \phantom{0}3.6 & 1.2 & \phantom{0}6.0 \\
        $\mathsf{medium}$ (\%) & \phantom{0}5.9 & 0.9 & \phantom{0}8.2\\
        $\mathsf{hard}$ (\%) & \phantom{0}8.0 & 1.5 & 10.3 \\
        $\mathsf{extra}$ (\%) & 11.0 & 3.5 & 17.6 \\
        \hline
        all data (\%) & \phantom{0}6.5 & 1.0 & \phantom{0}9.0\\
    \end{tabular}
    \caption{The false positive rate of single denotation accuracy (i.e., checking denotation only on the databases originally released in \citet{yu2018spider}) for each difficulty split. We report the mean / std / max of these two statistics among 21 dev set submissions.}
    \label{tab:orig-exec-fp}
\end{table}

\subsection{Errors of Traditional Metrics}\label{errors}
Given that test suite evaluation empirically provides an improved approximation of semantic equivalence, we use test suite accuracy as ground truth and retrospectively examine how well ESM approximates semantic accuracy.
We calculate the false positive/false negative rate for each difficulty split and report the mean, standard deviation, and max for all 21 submissions.

Table \ref{tab:fpfn} shows the results. 
ESM leads to a nontrivial false negative rate of $2.6\%$ on average, and $8.1\%$ in the worst case.
The error becomes larger for harder fractions of queries characterized by more complex queries. 
On the $\mathsf{hard}$ fraction, false negative rate increases to $4\%$ on average and $12.1\%$ in the worst case.

In Table \ref{tab:orig-exec-fp}, we report the difference between test suite accuracy and single denotation accuracy, which effectively means checking denotations of the model-predicted queries only on the databases from the original dataset release \cite{yu2018spider}.
In the worst case, single denotation accuracy creates a false positive rate of 8\% on the entire development set, and 4\% more on the $\mathsf{extra hard}$ fraction.

\subsection{Correlation with Existing Metrics} \label{sec:metric-corr}
Could surface-form based metric like ESM reliably track improvements in semantic accuracy?
\begin{figure}[h]
    \begin{subfigure}{.24\textwidth}%
    \label{fig:all-corr}%
    \includegraphics[width=\linewidth]{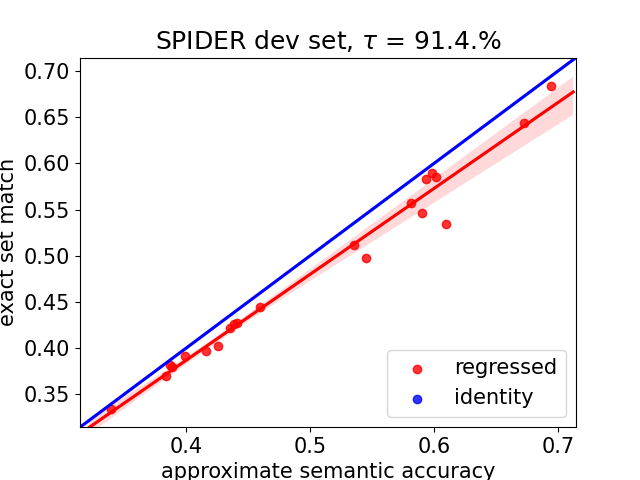}%
    \caption{$\tau = 91.4\%$ on all \\queries in the dev set.}
    \end{subfigure}%
    \begin{subfigure}{.24\textwidth}%
      \includegraphics[width=\linewidth]{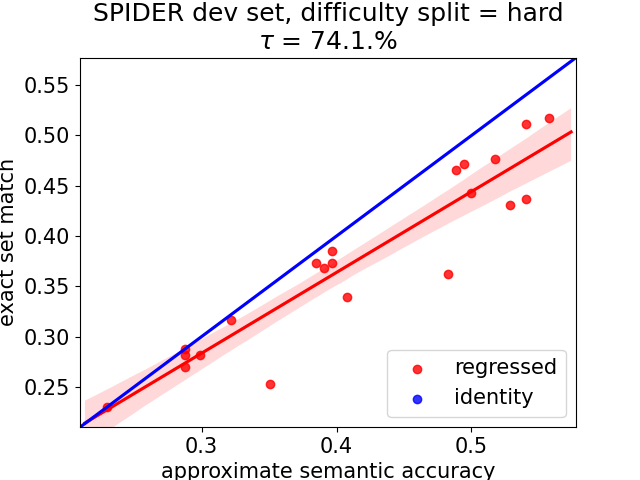}%
      \caption{$\tau = 74.1\%$ on \\$\mathsf{hard}$ fraction of the dev set.}
    \label{fig:extra-corr}%
    \end{subfigure}%
    \caption{Kendall $\tau$ correlation between exact set match and test suite accuracy. 
    Each dot is a dev set submission to the \spider{} leaderboard.
    }%
    \label{fig:m-corr}
\end{figure}
\begin{figure}[h]
    \begin{subfigure}{.24\textwidth}%
    \label{fig:orig-exec-all-corr}%
    \includegraphics[width=\linewidth]{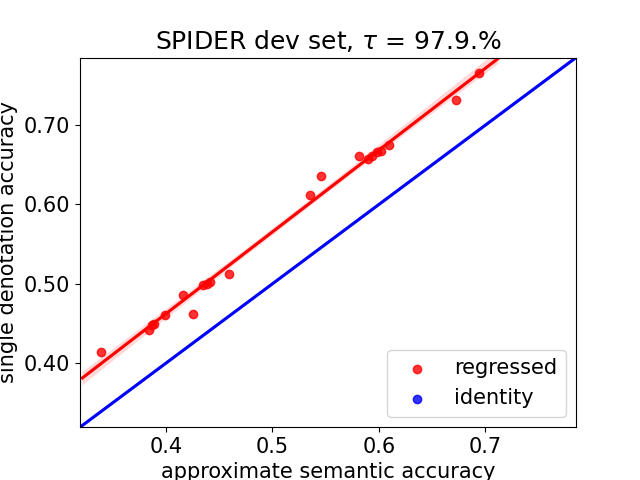}%
    \caption{$\tau = 97.9\%$ on all \\queries in the dev set.}
    \end{subfigure}%
    \begin{subfigure}{.24\textwidth}%
      \includegraphics[width=\linewidth]{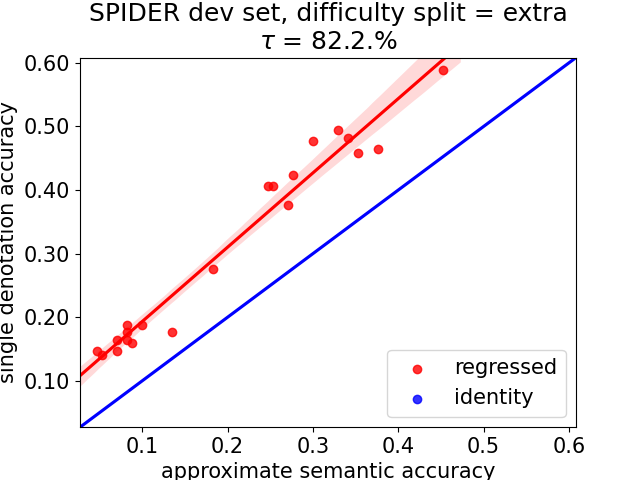}%
      \caption{$\tau = 82.2\%$ on $\mathsf{extra hard}$\\ fraction of the dev set.}
    \label{fig:orig-exec-extra-corr}%
    \end{subfigure}%
    \caption{Kendall $\tau$ correlation between single execution accuracy as originally defined in \citet{yu2018spider} and test suite accuracy. 
    Each dot is a dev set submission to the \spider{} leaderboard.
    }%
    \label{fig:exec-corr}
\end{figure}
We plot ESM against test suite accuracy for all 21 dev set submissions in Figure \ref{fig:m-corr}. 
On a macro level, ESM correlates well with test suite accuracy with Kendall $\tau$ correlation 91.4\% in aggregate;
however, the correlation decreases to 74.1\% on the $\mathsf{hard}$ fraction. 
Additionally, ESM and test suite accuracy starts to diverge as model performance increases.
These two facts jointly imply that as models are becoming better at harder queries, ESM is no longer sufficient to approximate semantic accuracy. 
On a micro level, when two models have close performances, improvements in semantic accuracy might not be reflected by increases in ESM. 
On the $\mathsf{hard}$ fraction, 5 out of 21 submissions have more than four others that have lower test suite accuracy but higher ESM scores
(i.e., there are five dots in Figure \ref{fig:extra-corr} such that for each of them,  four other dots is located in its upper left).

Figure \ref{fig:exec-corr} plots the correlation between single denotation accuracy against test suite accuracy.
On the $\mathsf{extra hard}$ fraction, four submissions have more than three others that have higher single denotation accuracy but lower test suite accuracy.
Checking denotation only on the original database is insufficient.

We list the Kendall $\tau$ correlations between test suite accuracy and different metrics in Table \ref{tab:all-metrics-corr} and plot them in the appendix Section \ref{all-corr-plots}.
The correlation with the current official metric is low without fixing the issue (3) identified in Section \ref{adapted-metric}. 

\begin{table}[h]
    \centering
    \scalebox{0.94}{
    \begin{tabular}{cccc }
    Difficulty & Adapted & Official & Single Denot. \\
    \hline
    $\mathsf{easy}$ (\%) & 91 & 86 & 90 \\
    $\mathsf{medium}$ (\%) & 90 & 37 & 96 \\
    $\mathsf{hard}$ (\%) & 75 & 28 & 94 \\
    $\mathsf{extra}$ (\%) & 91 & 20 & 82 \\
    \hline
    all data (\%) & 91 & 40 & 98 \\
    \end{tabular}
    }
    \caption{Kendall $\tau$ correlation between various metrics and test suite accuracy across 21 submissions.
    \textbf{Adapted} refers to ESM after we fix the issue (3) identified in Section \ref{adapted-metric}. \textbf{Official} refers to directly running the official evaluation script to evaluate, and \textbf{Single Denot.} refers to only checking denotation on the one database provided by \citet{yu2018spider}.}
    \label{tab:all-metrics-corr}
\end{table}

\subsection{Computational Efficiency} \label{sec:efficiency}
On average, we distill 42 databases for each of the 1034 queries.
In total, there are 695 databases since queries with the same database schema share the same test suite.
These databases take 3.27GB in space (databases from the original datasets take 100.7MB). 
Running the gold queries on the entire test suite takes 75.3 minutes on a single CPU (compared to 1.2 minutes on the databases from the original datasets).
Although test suite evaluation consumes more space and computational resources than single denotation evaluation, it is parallelizable and affordable by most researchers. 

We may speed up the evaluation by checking denotation only on a single database sampled from the distribution $\mathcal{I}_{g}$.
While this sped-up version sacrifices precision for speed,
\textit{retrospectively}, it produces the exact same outcomes as running the full test suite on the 21 submissions.
Therefore, the sped-up version might be useful when occasional errors are tolerable (e.g. denotation based training).
However, we still recommend using the full test suite for reliable evaluation, since a single sample from $\mathcal{I}_{g}$ cannot distinguish all neighbors, and checking denotations on multiple databases with comprehensive code coverage is always more reliable, especially when we have no prior information about the model-predicted queries.

\begin{figure*}
    \centering
    \includegraphics[scale=0.60]{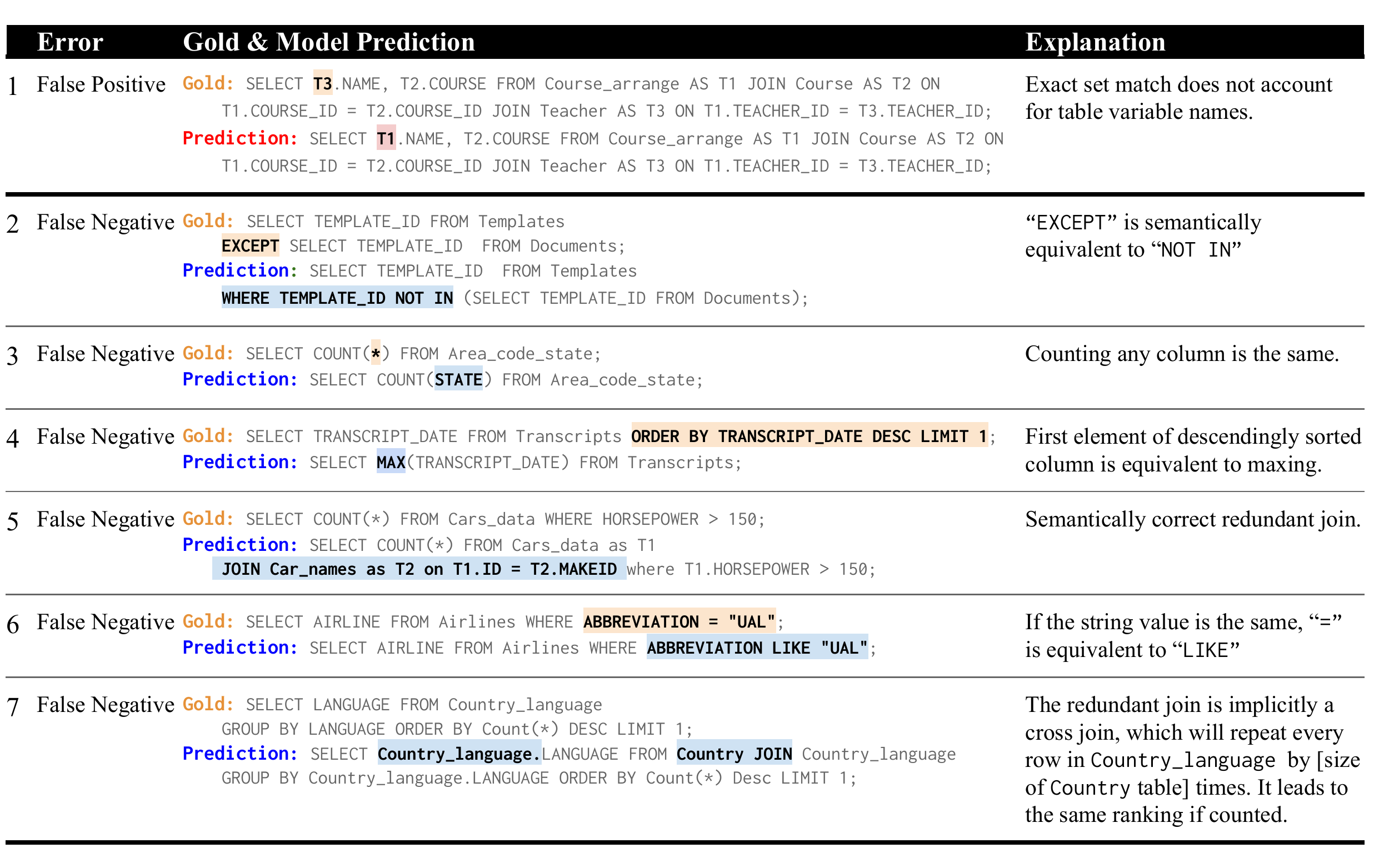}
    \caption{Representative examples where the exact set match (ESM) metric is different from test suite accuracy.
    False Positives happen when ESM judges a model-predicted query to be correct but test suite accuracy judges it to be wrong.
    False Negatives happen when the reverse takes place.
    }
    \label{fig:fpfn-analysis}
    \vspace{-2mm}
\end{figure*}

\section{Metrics Comparison and Analysis} \label{anlaysis-all}

We explain how ESM and test suite accuracy differ and provide representative examples (Figure \ref{fig:fpfn-analysis}). 

\paragraph{False Positives}
Although standard ESM is strict, the adapted ESM (Section \ref{adapted-metric}) can introduce false positives because it ignores table variable names. 
See Figure \ref{fig:fpfn-analysis} row 1 for an example. 

\paragraph{False Negatives}
Row 2-4 shows that slightly complicated queries usually have semantically equivalent variants, and it is nontrivial to tell whether they are semantically equivalent unless we execute them on a test suite or manually verify.

Nevertheless, even though test suite accuracy reliably approximates semantic accuracy according to our observation, researchers might also care about other aspects of a model-predicted query. 
Semantic accuracy is only concerned with \textit{what} are the denotations of a query, but not \textit{how} it calculates them. 
For example, Figure \ref{fig:fpfn-analysis} row 5 represents one of the most common types of false negatives, where the model-predicted query chooses to join other tables even though it is unnecessary. 
While semantically correct, the model-predicted query increases running time. 
Figure \ref{fig:fpfn-analysis} row 7 exhibits a similar but more complicated and rare example. 

Inserting gold values into model-predicted queries as described in Section \ref{sec:data} might also unexpectedly loosen the semantic accuracy metric.
For example, in Figure \ref{fig:fpfn-analysis} row 6, the model-predicted query uses the ``\texttt{LIKE}" keyword rather than the ``\texttt{=}" operator.
By SQL style conventions, ``\texttt{LIKE}" usually precedes a value of the form ``\texttt{\%[name]\%}" and corresponds to natural language query ``\texttt{contains [name]}" rather than ``\texttt{matches [name]}";
it seems plausible that the model does not understand the natural language query.
However, if we replace the wrong value ``\texttt{\%[name]\%}" with the gold value ``\texttt{[name]}" after the ``\texttt{LIKE}" operator,
the predicate becomes semantically equivalent to ``\texttt{= [value]}" and hence makes the query semantically correct. 
Value prediction is a crucial part of evaluating Text-to-SQL models.

\section{Discussion and Conclusion}
\paragraph{Semantic Evaluation via Test Suites} 
We propose test suite accuracy to approximate the semantic accuracy of a Text-to-SQL model, by automatically distilling a small test suite with comprehensive code coverage from a large number of random inputs.
We assure test suite quality by requiring it to distinguish neighbor queries and manually examining its judgments on model-predicted queries.
Our test suites will be released for eleven datasets so that future works can conveniently evaluate test suite accuracy. 
This metric better reflects semantic accuracy, and we hope that it can inspire novel model designs and training objectives.

Our framework for creating test suites is general and only has two requirements: (1) the input is strongly typed so that the fuzzing distribution $\mathcal{I}_{g}$ can be defined and the sample input can be meaningfully executed,
and (2) there exist neighbor queries $N_{g}$ that are semantically close but different from the gold $g$.
Since these two conditions hold in many execution environments, our framework might potentially be applied to other logical forms, such as $\lambda$-DCS \cite{liang2013lambda}, knowledge graphs \cite{lin-etal-2018-multi-hop}, and python code snippets \cite{yin2018mining, oda2015learning} if variable types can be heuristically extracted.
We hope to see more future work that evaluates approximate semantic accuracy on the existing benchmarks and formulates new tasks amenable to test suite evaluation.

We do not attempt to solve SQL equivalence testing in general. 
While our test suite achieves comprehensive code coverage of the gold query, it might not cover all the branches of model-predicted queries. 
Adversarially, we can always construct a query that differs from the gold only under extreme cases and fools our metric.
Fortunately, we never observe models making such pathological mistakes. 
However, it is necessary to revisit and verify this hypothesis some time later due to Goodhardt's law, since researchers will optimize over our metric.

\paragraph{Beyond Semantic Evaluation}
Although test suite evaluation provably never creates false negatives in a strict programming language sense, it might still consider ``acceptable answers" to be wrong and result in false negatives in a broader sense. 
For example, in a database of basketball game results, the predicate ``\texttt{A\_wins}" is equivalent to ``\texttt{scoreA $>$ scoreB}" according to common sense. 
However, such a relation is not explicitly reflected in the database schema, and our procedure might generate an ``unnatural" database where ``\texttt{scoreA $>$ scoreB}" but not ``\texttt{A\_wins}", hence distinguishing the model-predicted query from the gold. 
Fortunately, this issue is mitigated by current models. 
If ``\texttt{A\_wins}"  is mentioned in the text, the model would prefer predicting ``\texttt{A\_wins}"  rather than ``\texttt{scoreA $>$ scoreB}". 
Nevertheless, to completely solve this issue, we recommend future dataset builders to explicitly define the database generation procedure. 
Automatic constraint induction from database content and schema descriptions might also be possible, which is on its own an open research problem.

Additionally, some answers might be pragmatically acceptable but semantically wrong.
For example, if a user asks ``who is the oldest person?", the correct answer is a person's name. 
However, it also makes sense to return both the name and age columns, with the age column sorted in descending order.
Collecting multiple gold SQL query references for evaluation (like machine translation) might be a potential solution.

Finally, as discussed in Section \ref{anlaysis-all}, there might be other crucial aspects of a model-predicted query beyond semantic correctness.
Depending on the goal of the evaluation, other metrics such as memory/time efficiency and readability are also desirable and complementary to test suite accuracy. 

\section*{Acknowledgement}
We thank Wanyong Feng, Naihao Deng and Songhe Wang for rewriting queries and cleaning databases from \citet{finegan-dollak-etal-2018-improving}.
We thank Rishabh Agarwal, Tong Guo, Wonseok Hwang, Qin Lyu, Bryan McCann, Chen Liang, Sewon Min, Tianze Shi, Bailin Wang and Victor Zhong for responding to our cold email looking for WikiSQL model-predicted queries.

\bibliography{anthology,emnlp2020}
\bibliographystyle{acl_natbib}
\appendix
\newpage \phantom{0} \newpage
\section{Appendix}

\subsection{Algorithmic Description of Section \ref{sec:select}} \label{algo-description}
\begin{algorithm}
\SetAlgoLined
$S_{g} := \emptyset, N := N_{g}$ \;
\For{$t = 1, 2, \dots 1000$}
    {$w_{t} \sim \mathcal{I}_{g}$\;
    \For{$q \in N_{g}$}
    {\uIf{$D_{\{w_{t}\}}(q, g) = 1\;$}{
    $S_{g}.add(w_{t})$\;
    $N.remove(q)\;$}}
}
\Return{$S_{g}\;$}
\caption{Distilling a test suite $S_{g}$. $N_{g}$ is the set of neighbor queries of $g$; $\mathcal{I}_{g}$ is a distribution of database inputs.}
\label{construct-S}
\end{algorithm}

\subsection{Test Suite for Other Datasets} \label{sec:otherstats}
\paragraph{Data} We download Academic, Advising, ATIS, Geography, IMDB, Restaurants, Scholar and Yelp from \citet{finegan2018improving}.
For each dataset, we distill a test suite for the test split if it is already defined; otherwise we distill for the entire dataset.

We distill one shared test suite for the development set of \spider{} \cite{yu2018spider}, \cosql{} \cite{yu-etal-2019-cosql} and \sparc{} \cite{yu-etal-2019-sparc}, since they share the same 20 database schemata.

\paragraph{Test Suite Statistics}
The detailed test suite statistics can be seen in Table \ref{tab:testsuitestats}. 
The following list describes what each column represents:

\begin{itemize}
    \item \# Queries: the number of SQL queries we generate test suites for. Notice that this is not the full dataset size.
    \item Time: the time needed (in minute) to execute all the gold queries on its corresponding test suite on a single CPU. The smaller the value, the better.
    \item Size: the total size (in Giga-Bytes(G)/Mega-Bytes(M)) of the test suite. The smaller the value, the better.
    \item OrigSize: the size of the databases (in Giga-Bytes) in the original release. 
    \item $|N_{g}|$: the average number of neighbor queries generated for each gold query $g$ in the dataset. 
    \item Undistinguished: the fraction of neighbor queries that cannot be distinguished by the test suite. The smaller the value, the better.
    \item \# ``Reliable": the estimated fraction of gold queries in a dataset that can be \emph{reliably evaluated} (defined below). The larger the value, the better.
    
\end{itemize}

\begin{table*}[]  
    \centering
    \begin{tabular}{cccccccc} 
        Dataset & \# Queries & Time $\downarrow$ & Size $\downarrow$ & OrigSize & $|N_{g}|$ & Undistinguished $\downarrow$ & \# ``Reliable" $\uparrow$\\ 
        \hline
        \spider{} & 1034 & 75.3m & 3.27G & 0.10G & \phantom{0}94 & 0.44\% & 100.0\%\\
        \hline
        \cosql{} & 1007 & 75.6m & 3.27G & 0.10G & \phantom{0}93 & 0.48\% & 100.0\%\\
        \hline
        \sparc{} & 1203 & 86.7m & 3.27G & 0.10G & \phantom{0}81 & 0.71\% & 100.0\%\\
        \hline
        Academic & \phantom{0}189 & \phantom{0}1.6m & 0.03G & 4.81G & 368& 1.36\% & \phantom{0}94.7\% \\
        \hline
        Advising & \phantom{00}76 & \phantom{0}1.7m & 0.14G & 0.03G & 520 & 0.91\% & \phantom{0}63.2\%\\
        \hline
        ATIS & \phantom{00}93 & 19.2m & 0.92G & 0.06G & 974 & 0.63\% & \phantom{0}76.3\%\\
        \hline
        Geography & \phantom{00}51 & \phantom{0}0.4m & 2.21M & 0.26M & 108 & 5.28 \% & \phantom{0}88.2\%\\
        \hline
        IMDB & \phantom{00}97 &  \phantom{0}0.8m & 0.02G & 0.99G & 253 & 0.23\% & 100.0\%\\
        \hline
        Restaurants & \phantom{00}23 & \phantom{0}0.2m & 1.37M & 1.03M & 379 & 0.14\% & 100.0\%\\
        \hline
        Scholar & \phantom{0}101& \phantom{0}0.9m & 9.43M & 6.45G & 107 & 0.54\% & 92.1\% \\
        \hline
        Yelp & \phantom{0}122 & \phantom{0}1.4m & 0.02G &  2.15G & 274 & 0.07\% & 98.3\% \\
        \hline
        
    \end{tabular}
    \caption{Detailed test suite statistics by datasets. Appendix Section \ref{sec:otherstats} includes detailed explanation of each column name.
    $\downarrow$/$\uparrow$ means that we hope the number to be small/large.
    \spider{}, \cosql{} and \sparc{} share the same test suite.
    }
    \label{tab:testsuitestats}
\end{table*}

\spider{}, \cosql{} and \sparc{} have approximately the same statistics, since they share the same database schema and annotation convention. 
The other eight datasets have significantly longer queries with much more ``\texttt{JOIN}" and ``\texttt{WHERE}" operations.
Hence, there are more spans to drop and more neighbors are generated per query.

\paragraph{Reliability}
Table \ref{tab:testsuitestats} column ``Undistinguished" implies that fuzzing cannot distinguish a non-trivial fraction of neighbor queries for some datasets.
Besides cases where the neighbor queries are accidentally semantically equivalent to the gold, there are two major categories where fuzzing fails to distinguish semantically inequivalent neighbors.
\begin{itemize}
    \item The gold query contains too many ``\texttt{WHERE}" operations.
    For example, among the 93 queries in the ATIS test split, the maximum number of ``\texttt{WHERE}" operations is 24 for a single query, whereas this number is only 2 among 1034 queries in the \spider{} development set.
    Distinguishing two queries that differ by only one ``\texttt{WHERE}" operation is hard because the randomly sampled database needs to have a row that exactly satisfies all the ``\texttt{WHERE}" clauses.
    \item The gold query contains predicates like ``\texttt{WHERE COUNT(*) > 5000}". 
    Distinguishing ``\texttt{WHERE COUNT(*) > 5000}" from ``\texttt{WHERE COUNT(*) > 4999}" requires the number of the target (intermediate) table to have a size exactly 5000.
    Such a requirement is particularly hard for randomly generated databases.
\end{itemize}

We say that a datapoint can be \emph{reliably evaluated} if all of its undistinguished neighbors do not fall into the above two categories; then we estimate the fraction of datapoints that can be reliably evaluated for each dataset in Table \ref{tab:testsuitestats}.
Fortunately, the majority of queries can be reliably evaluated for every dataset. 
Future manual efforts to hand-craft test suite might be needed to distinguish the neighbor queries and make test suite evaluation more reliable on ATIS and Advising.

Finally, \citet{49288} evaluates execution accuracy only on datapoints where the gold denotation is not empty.
In comparison, at least one database from our test suite produces non-empty gold denotation for every datapoint in all eleven datasets.

\subsection{Evaluation on WikiSQL} \label{validateonwikisql}

We show that our test suite evaluation strategy also works well for model-predicted queries on WikiSQL \cite{zhongSeq2SQL2017}.
The dev/test set contains 8420/15878 SQL queries, respectively.

\paragraph{Model-Predicted Queries}
We reached out to authors of individual works to obtain real model predictions on WikiSQL, and heard back from \citet{min-etal-2019-discrete, McCann2018TheNL, lyu2020hybrid, Hwang2019ACE, He2019XSQLRS, Shi2018IncSQLTI, guo-2017-deep, DBLP:conf/icml/AgarwalLS019, liang2018memory, wang-etal-2019-learning}.

We use the model-predicted queries from the first six works cited above since they provided model-predicted queries in the format consistent with \citet{zhongSeq2SQL2017}, which can be easily converted into SQL queries.
Specifically, we consider the model-predicted queries from the following eight models: MQAN unordered \cite{McCann2018TheNL}, X-SQL \cite{He2019XSQLRS}, HydraNet with/without Execution Guidance \cite{lyu2020hybrid}, IncSQL \cite{Shi2018IncSQLTI}, SQLova with/without Execution Guidance \cite{Hwang2019ACE} and HardEM \cite{min-etal-2019-discrete}.
This provides us in total (8420 + 15878) $\times $ 8 $\approx$ 200K model-predicted queries.

\paragraph{Test Suite Generation}
We run the fuzzing algorithm (Section \ref{sec:fuzzing}) as before to create test suite.
Since the most complex query in WikiSQL is simple and only consists of a single ``$\texttt{WHERE}$" clause with an aggregation operation, our test suite can distinguish all the neighbors. 

\paragraph{Metric Difference}
To check whether our test suite reliably evaluates semantic accuracy, we examine model-predicted queries where test suite accuracy disagrees with Exact Set Match (ESM) (as in Section \ref{reliability}).

We find that there is only one pattern where a semantically correct prediction is considered wrong by ESM: counting any column of a table gives exactly the same denotation. For example, ``\texttt{SELECT count(col1) from Table}" is semantically equivalent to ``\texttt{SELECT count(col2) from Table}" but different in surface form.
After implementing a rule to filter out this equivalence pattern, we only find one model-predicted query that is considered wrong by ESM but correct by test suite accuracy, and we present it below.

The gold query is 

\texttt{SELECT MAX(col2) table WHERE col4 = 10;}

, while the model-predicted query is  

\texttt{SELECT MAX(col2) FROM table WHERE col2 > 10 AND col4 = 10;}

.This model-predicted query is not semantically equivalent to the gold, and hence our test suite evaluation makes an error. 
It over-generates a clause ``\texttt{WHERE col2 > 10}" that is not covered by the test suite.
None of our sampled database leads to a gold denotation fewer or equal to 10, which is a necessary and sufficient condition to distinguish these two queries.

To conclude, on WikiSQL, we only find 1 out of 200K model-predicted queries where our test suite accuracy makes an error, while we are able to verify that our test suite accuracy is correct for all the other model-predicted queries. 
This further implies that our method to develop semantic evaluation is robust and not dataset-specific.

On the other hand, however, the only semantically equivalent variant of the gold query in WikiSQL is replacing the column to be counted.
Since we might still want to check which column the model-predicted query is counting for code readability, we do NOT recommend researchers to use test suite accuracy for WikiSQL.

\subsection{Correlation Plot with Other Metrics}\label{all-corr-plots}

\begin{figure}[h]
    \begin{subfigure}{.24\textwidth}%
    \includegraphics[width=\linewidth]{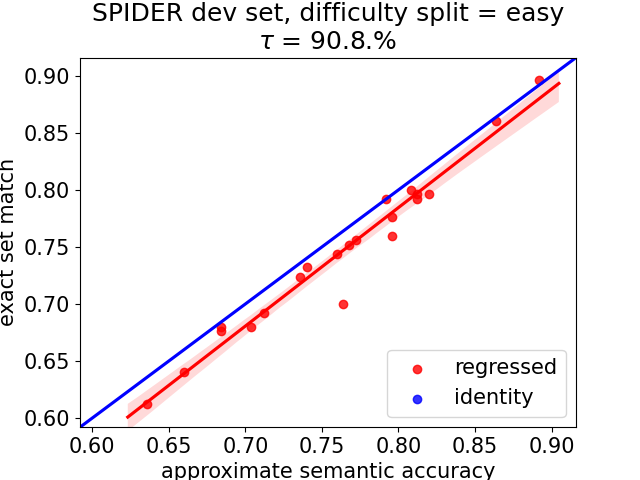}%
    \caption{$\tau = 90.8\%$ on \\easy fraction.}
    \end{subfigure}%
    \begin{subfigure}{.24\textwidth}%
    \includegraphics[width=\linewidth]{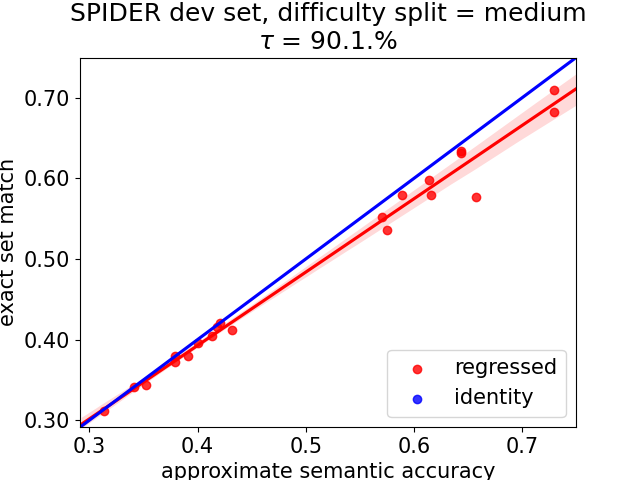}%
    \caption{$\tau = 90.1\%$ on \\medium fraction.}
    \end{subfigure}\\%
    \begin{subfigure}{.24\textwidth}%
      \includegraphics[width=\linewidth]{figures/fuzzing-based_accuracyexact_set_matchhard.png}%
      \caption{$\tau = 74.1\%$ on \\hard fraction.}
    \end{subfigure}%
    \begin{subfigure}{.24\textwidth}%
      \includegraphics[width=\linewidth]{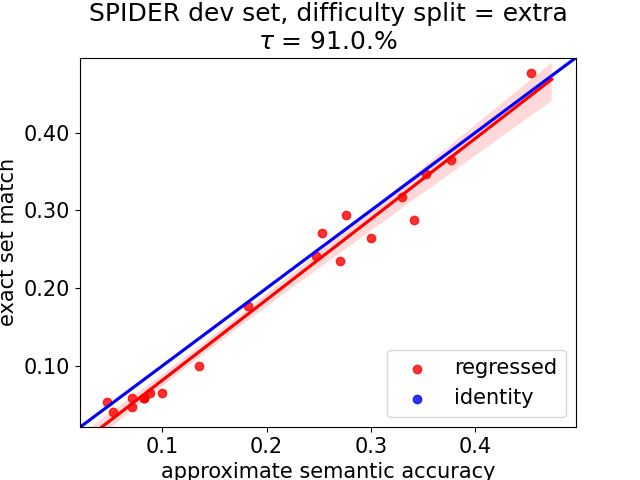}%
      \caption{$\tau = 91.0\%$ on \\extra hard fraction.}
    \end{subfigure}\\%
    \begin{subfigure}{.24\textwidth}%
      \includegraphics[width=\linewidth]{figures/fuzzing-based_accuracyexact_set_matchall.png}%
      \caption{$\tau = 91.4\%$ on \\all data.}
    \end{subfigure}%
    \caption{Kendall $\tau$ correlation between \textbf{adapted  exact set match} and fuzzing-based accuracy. 
    Each dot in the plot represents a dev set submission to the \spider{} leader board.
    }%
    \label{fig:all-em-corr}
    
\end{figure}

\begin{figure}[h]
    \begin{subfigure}{.24\textwidth}%
    \includegraphics[width=\linewidth]{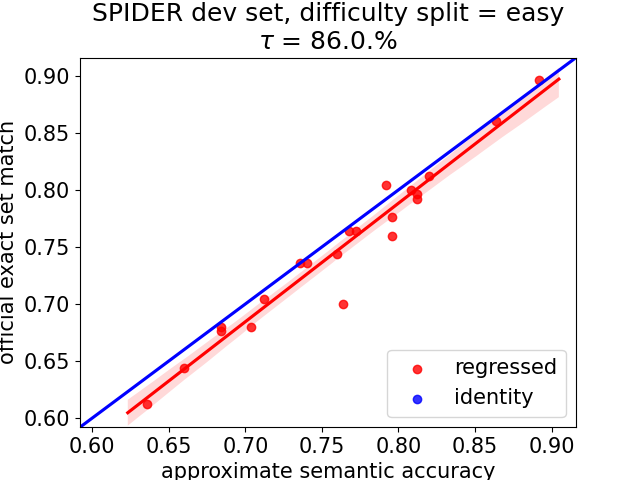}%
    \caption{$\tau = 86.0\%$ on \\easy fraction.}
    \end{subfigure}%
    \begin{subfigure}{.24\textwidth}%
    \includegraphics[width=\linewidth]{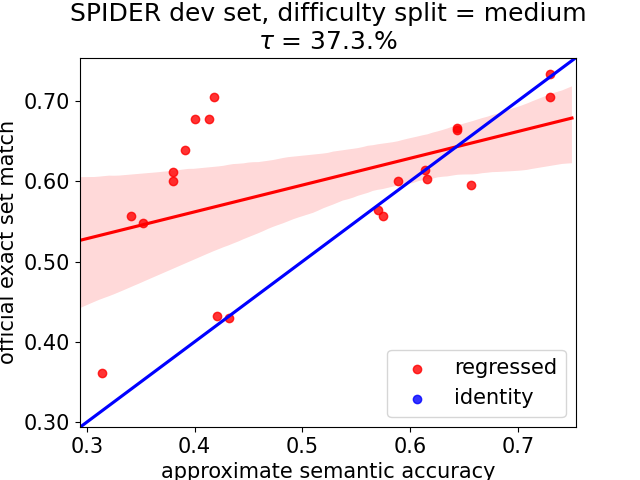}%
    \caption{$\tau = 37.3\%$ on \\medium fraction.}
    \end{subfigure}\\%
    \begin{subfigure}{.24\textwidth}%
      \includegraphics[width=\linewidth]{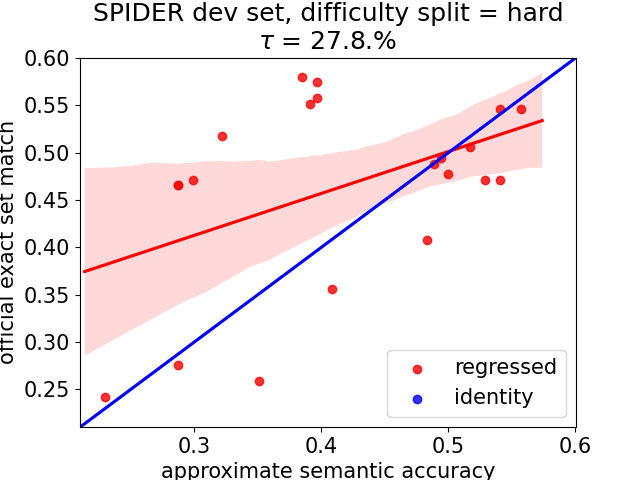}%
      \caption{$\tau = 27.8\%$ on \\hard fraction.}
    \end{subfigure}%
    \begin{subfigure}{.24\textwidth}%
      \includegraphics[width=\linewidth]{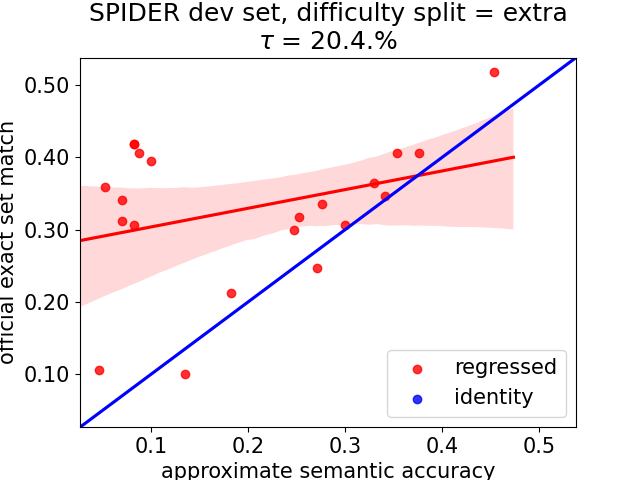}%
      \caption{$\tau = 20.4\%$ on \\extra hard fraction.}
    \end{subfigure}\\%
    \begin{subfigure}{.24\textwidth}%
      \includegraphics[width=\linewidth]{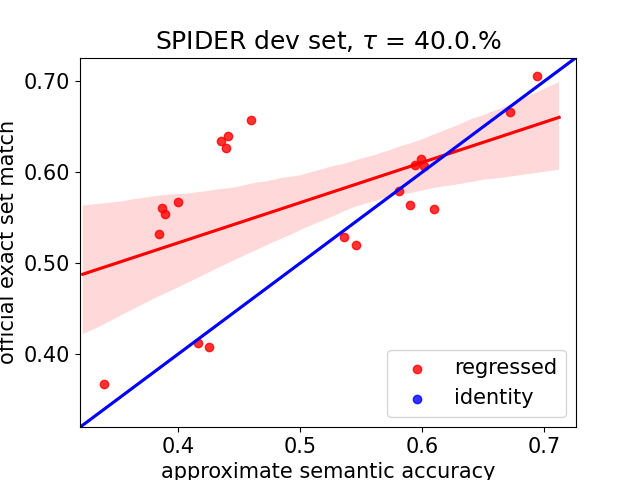}%
      \caption{$\tau = 40.0\%$ on \\all data.}
    \end{subfigure}%
    \caption{Kendall $\tau$ correlation between the official \textbf{\spider{} exact set match} and fuzzing-based accuracy. 
    Each dot in the plot represents a dev set submission to the \spider{} leader board.
    }%
    \label{fig:all-official-corr}
    
\end{figure}

\begin{figure}[h]
    \begin{subfigure}{.24\textwidth}%
    \includegraphics[width=\linewidth]{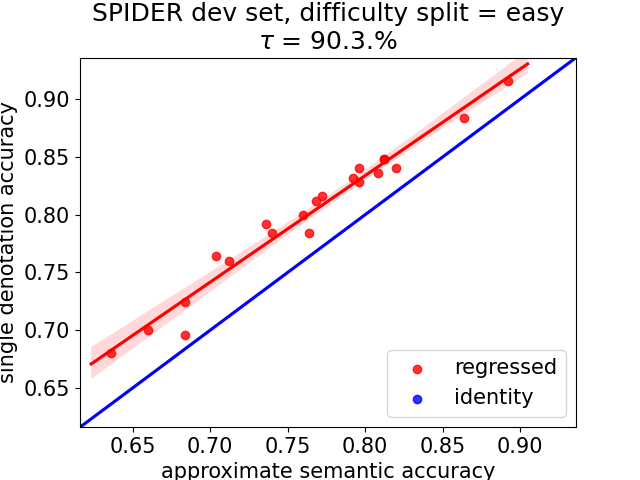}%
    \caption{$\tau = 90.3\%$ on \\easy fraction.}
    \end{subfigure}%
    \begin{subfigure}{.24\textwidth}%
    \includegraphics[width=\linewidth]{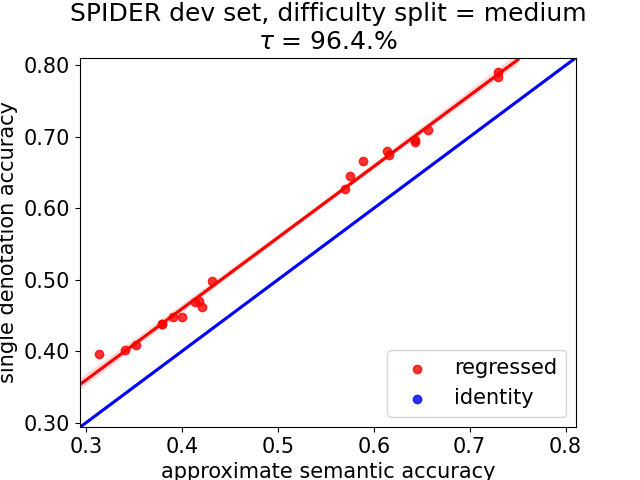}%
    \caption{$\tau = 96.4\%$ on \\medium fraction.}
    \end{subfigure}\\%
    \begin{subfigure}{.24\textwidth}%
      \includegraphics[width=\linewidth]{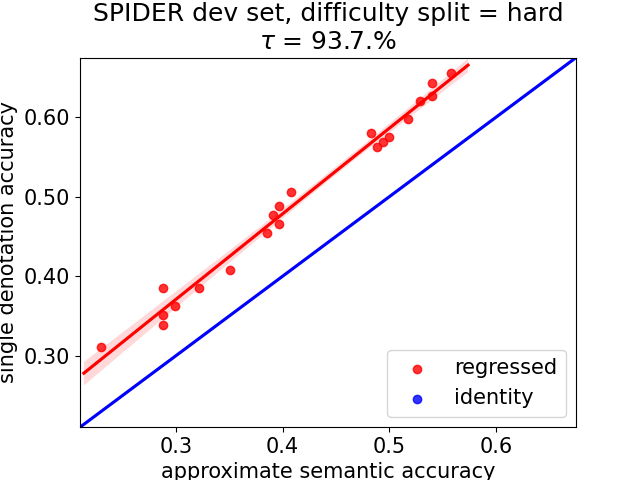}%
      \caption{$\tau = 93.7\%$ on \\hard fraction.}
    \end{subfigure}%
    \begin{subfigure}{.24\textwidth}%
      \includegraphics[width=\linewidth]{figures/fuzzing-based_accuracysingle_denotation_accuracyextra.png}%
      \caption{$\tau = 82.2\%$ on \\extra hard fraction.}
    \end{subfigure}\\%
    \begin{subfigure}{.24\textwidth}%
      \includegraphics[width=\linewidth]{figures/fuzzing-based_accuracysingle_denotation_accuracyall.png}%
      \caption{$\tau = 97.9\%$ on \\all data.}
    \end{subfigure}%
    \caption{Kendall $\tau$ correlation between\textbf{ single denotation accuracy} and fuzzing-based accuracy. 
    Each dot in the plot represents a dev set submission to the \spider{} leader board.
    }%
    \label{fig:all-se-corr}
    
\end{figure}

We plot the correlation between test suite accuracy and (1) adapted exact set match (Figure \ref{fig:all-em-corr}), (2) official \spider{} exact set match (Figure \ref{fig:all-official-corr}), and (3) single denotation accuracy (Figure \ref{fig:all-se-corr}) on each fraction of the difficulty split.

\end{document}